\documentclass[runningheads]{llncs}
\usepackage{graphicx}
\usepackage{amsmath,amssymb,amsfonts}
\usepackage{multirow, array}
\usepackage{cite}
\usepackage[T1]{fontenc}
\begin{document}
\title{Core and Periphery as Closed-System Precepts for Engineering General Intelligence}
%
%
\author{Tyler Cody$^{a}$, Niloofar Shadab$^{b}$, Alejandro Salado$^{c}$, Peter Beling$^{a, b}$}
\authorrunning{T. Cody et al.}
\titlerunning{Core and Periphery as Closed-System Precepts}
%
\institute{$^{a}$National Security Institute, Virginia Tech \\
$^{b}$Grado Department of Industrial and Systems Engineering, Virginia Tech \\
$^{c}$Department of Systems and Industrial Engineering, The University of Arizona \\
}
\maketitle              
\begin{abstract}

Engineering methods are centered around traditional notions of decomposition and recomposition that rely on partitioning the inputs and outputs of components to allow for component-level properties to hold after their composition. In artificial intelligence (AI), however, systems are often expected to influence their environments, and, by way of their environments, to influence themselves. Thus, it is unclear if an AI system's inputs will be independent of its outputs, and, therefore, if AI systems can be treated as traditional components. This paper posits that engineering general intelligence requires new general systems precepts, termed the core and periphery, and explores their theoretical uses. The new precepts are elaborated using abstract systems theory and the Law of Requisite Variety. By using the presented material, engineers can better understand the general character of regulating the outcomes of AI to achieve stakeholder needs and how the general systems nature of embodiment challenges traditional engineering practice.

\end{abstract}

\section{Introduction}

Engineering methods are still centered around traditional engineering notions of decomposing stakeholder needs and outcomes into component-level functions and recomposing those component-level functions into subsystems and systems \cite{salado2021systems}. This traditional approach of engineering by aggregation relies on a partitioning of inputs and outputs to allow for component-level properties to hold after composition \cite{wymore1967mathematical}. 
While the artificial intelligence (AI) in AI-enabled systems---systems with AI components or subsystems---may be well-treated as an individually addressable part at conception, the boundaries between an AI part, other aspects of its greater system, and the environment it interacts with face dissolution as the three intertwine.

This paper posits that whereas traditional engineering is driven by a focus on open systems and correspondingly on precepts of decomposition and recomposition, engineering general intelligence requires an alternative treatment and new precepts. This paper substantiates discourse on a new framework for engineering by challenging the legitimacy of existing precepts. Moreover, this paper proposes two new general systems precepts, \emph{core} and \emph{periphery}, and discusses their use. While previous work on the topic of embodiment explores related concepts \cite{shapiro2019embodied}, importantly, it does not explore embodiment as a consequence of general systems theory or directly identify the challenges to traditional engineering that embodiment presents. Using the material presented herein, engineers can better understand the general nature of regulating the outcomes of AI-enabled systems, and thereby of achieving stakeholder needs.

This paper is structured as follows. Embodied cognition is reviewed as a related, although differently motivated field of research in cognitive science. Then, the limitations of existing engineering practice are outlined. Discussion is lead to a review of the Law of Requisite Variety \cite{ashby1991requisite}, which is used as a basis to define core and periphery as closed-system precepts, and to explore their use in modeling AI. Before concluding, remarks are made on relevance.

\section{Related Work}

Embodied cognition is a cognitive science that considers the role body and environment, in addition to mind, play in cognitive processes, and, moreover, emphasizes a lack of distinction between the three \cite{pfeifer2004embodied}. Embodied cognition can be characterized as: ``a research program with no clear defining features other than the tenet that computational cognitive science has failed to appreciate the body's significance in cognitive processing and to do so requires a dramatic re-conceptualization of the nature of cognition and how it must be investigated'' \cite{shapiro2019embodied}.

Notions of embodiment are closely related to ecological psychology, which, eschewing the notion of cognition as computation, posits that cognitive processes, like perceptual processes, involve the whole organism as it moves through the environment \cite{michaels2014ten}. This contrasts the traditional view of computational cognitive scientists that cognitive processes require inference from ``impoverished'' inputs, which, on their face, do not contain enough information to solve problems, and therefore necessitate a kind-of Bayesian conditioning of inputs with background knowledge \cite{chomsky2006cognitive}. 

Attempts to bring embodied cognition from philosophy to the real-world include robotics and the use of dynamical systems. Embodied cognitive robotics limits, discredits, or otherwise avoids the use of internal representations and the use of symbolic logic over them---in the extreme, linking perception directly to action \cite{brooks1991new,brooks1991intelligence}. Some critics are quick to point out the subjectivity of determining what is and what is not a representation \cite{chemero2013radical,hutto2013radicalizing}, e.g., as sensors already bias inputs away from reality \cite{rydehn2021grounding}. Other critics strongly challenge scalability \cite{matthen2014debunking}.

Dynamical cognitive science treats cognition as a dynamical system: a continuous time relationship among the component sets of a system and the relations between them \cite{beer2000dynamical}. In essence, it favors the view of ``mind as a continuous event'' in the stead of ``mind as computer'' \cite{spivey2008continuity}. But real-world examples remain simple \cite{haken1985theoretical,smith1993dynamic,thelen2001dynamics,beer2003dynamics}, because dynamical systems quickly become complex and adaptive as they scale in intelligence \cite{weinbaum2017open}, thereby limiting scientific investigation. Although taking a formal systems view, dynamical cognitive science has fallen short of defining formal, general engineering precepts for intelligence.

Most often, embodied cognition is a topic of natural intelligence, and less so of AI, because AI is largely concerned with computation, and, in present day, with computational approaches to \emph{problem} solving tasks \cite{thorisson2016artificial}, as opposed to (cognitive) \emph{systems} which solve problems or come to be able to solve problems. As a result, questions regarding where cognition resides or where problem solving takes place are generally not within the scope of discourse \cite{cody2021mesarovician}. As such, in computer science, there are disparate research efforts with a broadened scope \cite{schick2010bodies, steels2018artificial}, but they often rely on their disparate specifics. This manuscript works outward from a general systems perspective, as opposed to from a cognitive psychologist or computer science perspective, to suggest new precepts for engineering embodiment.


\section{Existing Precepts and Their Limits}

Precepts for engineering AI must presume something of the nature of intelligence. There is ongoing research into defining intelligence and the properties it exhibits in engineered systems \cite{wang2019defining}. Some advocate that intelligence is measured by integrating a complexity-weighted performance measure over a set of tasks \cite{chollet2019measure}. Others advocate that intelligence is manifested as a minimization of complexity in state dynamics \cite{friston2010free}. And others yet still measure intelligence in terms of adversarial sequence prediction \cite{hibbard2011measuring, AlexanderHibbard+2021+1+25}. Each alternative definition leads the discussion of engineering intelligence in a different direction. This paper avoids the constricting effect of pursuing a specific definition of intelligence on the generality of results by focusing on precepts for the case when \emph{intelligence is a property of the relation between an system and its environment}---rather than a property of the system itself.

\begin{figure*}[t]
    \centering
    \includegraphics[width=.95\textwidth]{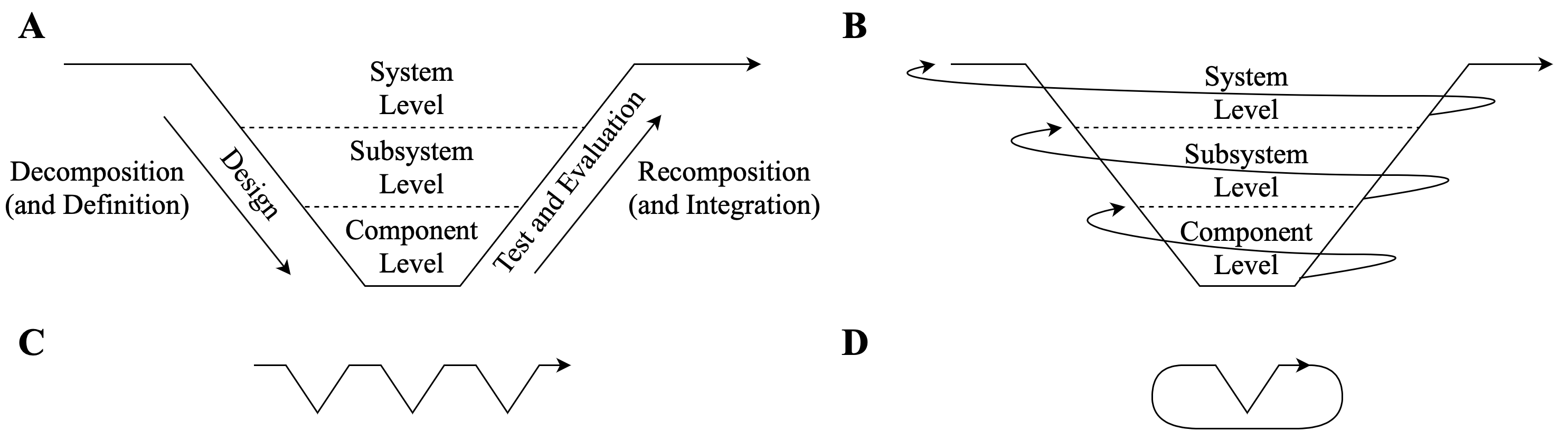}
    \caption{(A) depicts the traditional engineering process of decomposing systems into individually addressable components and recomposing them into systems. But, can AI-enabled systems with inter- and intra-level coupling as depicted by spiraling, multi-level dependencies in (B) be treated with the same precepts of decomposition and recomposition? The basic concept of engineering operations offered by traditional approaches follows the iterative decomposition and recomposition shown in (C). In contrast, the posited, highly coupling effect of intelligence imposes a continous concept of operations shown in (D).}
    \label{fig:v}
\end{figure*}

The latter case, that of intelligence as a property of a system, suggests a continuation of existing engineering practice. Given a system and needed outcomes of that system, systems engineers decompose the system into subsystems and their components, specify functional requirements on the components, and then distribute the engineering of each functional component to their respective disciplines. Subsequently, component-level solutions are recomposed into subsystems and, in turn, into the system as a whole, performing test and evaluation along the way, as shown in Figure \ref{fig:v}. Once properly composed, the system is deployed into operation, putting engineers in a holding pattern until another iteration of the so-called engineering ``V'' is desired \cite{haskins2006systems, klatt2009perspectives}. This traditional practice of engineering by following the mantra, ``If the parts work, and the interfaces between the parts work, then the whole will work'', is rooted in precepts of \emph{decomposition} and \emph{recomposition} and is in direct conflict with the environmental coupling that this paper posits as the definitive feature of engineering general intelligence.

Deep neural networks (DNNs) are exemplary of this phenomena. DNNs can be specified as a composition of functions that pass information from layer to layer in a way that meets certain mathematical requirements. Specifications of DNNs using functional requirements are nearly the same for the enormous number of systems where DNNs are applied. However, apparently, the outcomes \emph{needed} from DNNs vary greatly between systems. Thus, there is an apparent gap between achieving the needed outcomes a stakeholder has for a DNN and the functional requirements of a DNN. Whereas embodied cognition views this gap as the result of a flawed \emph{philosophical} view of the mind as computation from which an inappropriate characterization of the relation between stimuli and cognition is derived, this paper views this gap as the result of a flawed \emph{mathematical}---and therefore formal---foundation of traditional engineering which undergirds the dogma of engineering by composition.

If engineers cannot rely on functional decomposition and recomposition as precepts, what can they rely on? First, AI engineers must admit that they cannot readily specify the needs and outcomes of stakeholders into low-level functions and requirements. That is, simply ensuring input-output relationships will not reliably generate desired outcomes as it has in the past. With a renewed focus on the primacy of outcomes to input-output relations, engineers must then turn to new precepts that do not rely on persistent boundaries across the various subsystems and levels of abstraction in systems.



\section{Outcomes and Requisite Variety}

In general systems theory, systems are (often) defined as a relation on sets. General systems theory is thus (often) concerned with general conditions of relations on sets. These can be categorical, topological, algebraic, etc., however, set theory alone can be illuminating of the character of any of those more specific concerns. In the mid-twentieth century, Ashby used a particular notion of variety to study homeostasis---the ability to maintain certain variables within tight bounds despite changing contexts---in biological systems \cite{ashby1961introduction, ashby1991requisite}, and he made a remarkably general discovery regarding the nature of outcomes termed the \emph{Law of Requisite Variety}.

Consider two systems $S$ and $S_E$ where $S:\mathcal{X} \to \mathcal{Y}$ and $S_E: \mathcal{X}_E \to \mathcal{Y}_E$. Without loss of generality term $S$ the system and $S_E$ the environment. Suppose $S$ is acting as a regulator of $S_E$. Let $\mathcal{X}_{E \setminus S} = \mathcal{X}_E \setminus \mathcal{Y}$ where $\setminus$ denotes set difference. In other words, inputs to the environment $\mathcal{X}_E = \mathcal{X}_{E \setminus S} \cup \mathcal{Y}$. Consider a set of outcomes $\mathcal{Z}$ with support over $\mathcal{X}_{E \setminus S} \times \mathcal{Y}$, i.e., $\mathcal{X}_{E \setminus S} \times \mathcal{Y} \to \mathcal{Z}$. This notion of outcomes is identical to payoff matrices used in game theory. Let $V_A$ be termed \emph{variety} and be the Shannon entropy of a finite set $A$, i.e., 
\begin{equation}
\label{eq:var}
    V_A = - \sum^{|A|}_{i} p_i \log_2 p_i,
\end{equation}
where $|A|$ denotes the cardinality of $A$ and $p_i$ the probability of the $i$th element of $A$. Variety describes the number of unique elements in a system. 


The Law of Requisite Variety states that for one system to be a stable regulator of another, the variety of the regulator's output must be greater than or equal to the variety of the regulated system's input. Formally put, consider that (from \cite{ashby1991requisite})
\begin{equation}
\label{eq:req}
    \min V_\mathcal{Z} = \max \{ V_{\mathcal{X}_{E \setminus S}} - V_{\mathcal{Y}}, 0 \}.
\end{equation}
The Law of Requisite Variety can be defined as follows.
\begin{definition}[Law of Requisite Variety]
The \emph{Law of Requisite Variety} states that given $V_{\mathcal{X}_{E \setminus S}}$, the minimum variety of outcomes $\min V_\mathcal{Z}$ only decreases if $V_\mathcal{Y}$ increases.
\end{definition}
Only if $V_{\mathcal{Y}} \geq V_{\mathcal{X}_{E \setminus S}}$, is it information theoretically possible to determine outcomes $\mathcal{Z}$, i.e., $\min V_\mathcal{Z} = 0$.

In summary, Equation \ref{eq:req} suggests that when the environment's input variety is not well-matched by the regulating system's output variety, the variety of the set of possible outcomes is necessarily large, and therefore the system will struggle to achieve precise outcomes. In the words of Ashby, system $S$'s ``capacity as a regulator cannot exceed its capacity as a channel for variety''\cite{ashby1991requisite}. 


\section{Core and Periphery}

Ashby considered system survival as dependent on bounding varieties \cite{ashby1961introduction}. Let bounded varieties be system varieties that are invariant and let unbounded varieties be system varieties that are \emph{not} invariant. Informally speaking, one could identify those structures that are core to the functioning of a system with bounded varieties, and one could identify those that are peripheral to such a core with unbounded varieties. In the following, using the Law of Requisite Variety, this paper presents the formalism necessary to establish core and periphery as precepts.

\subsection{Definition}

Let $S$ be a system $S \subset \times \{\mathcal{X}, \mathcal{Y}\}$ and let $\overline{S}$ denote the component sets of $S$, i.e., $\{\mathcal{X}, \mathcal{Y}\}$. Let $\mathcal{X}^t$ denote the input structure at time $t$, and so forth. Bounded and unbounded varieties are distinguished by measuring the variety of a system's residual change over time. Let $R$ denote this residual change, i.e.,
\begin{equation}
\label{eq:residual}
    R_{\overline{S}}^{t, t'} = \{\mathcal{X}^{t'} \setminus \mathcal{X}^t, \mathcal{Y}^{t'} \setminus \mathcal{Y}^t\}
\end{equation}
$R_{\overline{S}}^{t, t'}$ gives the residual change in system structure between time $t$ and $t'$. The core and periphery are defined as follows.
\begin{definition}[Core and Periphery]
Consider a system $S$ at time $t$ and at a later time $t'$. The \emph{core} of $S$ from $t$ to $t'$ is 
\begin{equation}
    \mathcal{C} = \overline{S} \setminus R_{\overline{S}}^{t, t'}
\end{equation}
The \emph{periphery} of $S$ from $t$ to $t'$ is 
\begin{equation}
    \mathcal{P} = R_{\overline{S}}^{t, t'}.
\end{equation}
\end{definition}
The core are those elements of $S$'s component sets that are identical at times $t$ and $t'$, and the periphery are those elements that are not.


\subsection{Core and Periphery as Precepts}

A number of immediate uses of core and periphery as precepts are now considered.

\subsubsection{Symmetry}

\begin{figure*}[t]
    \centering
    \includegraphics[width=.80\textwidth]{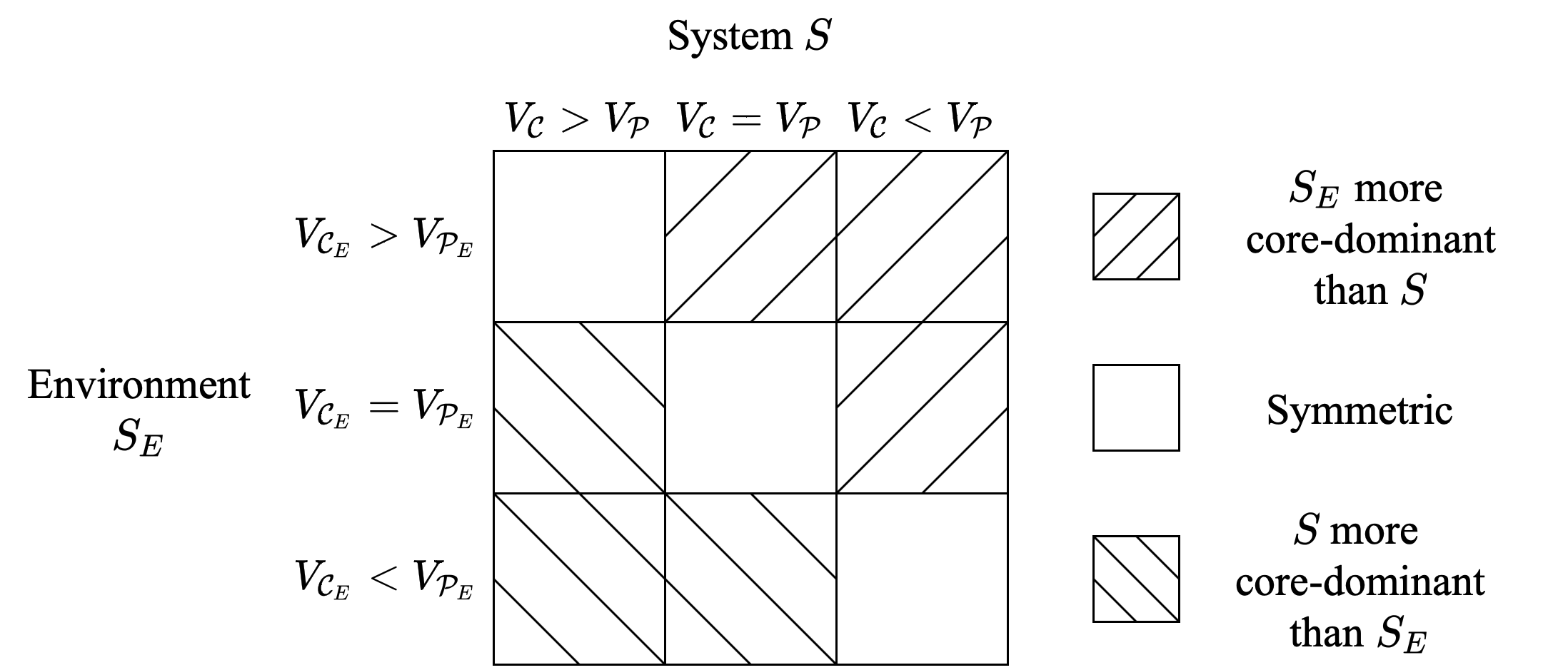}
    \caption{A table depicting possible symmetries and asymmetries between the core and periphery of the system and environment.}
    \label{fig:sym}
\end{figure*}

Consider that the environment $S_E$ has a core $\mathcal{C}_E$ and periphery $\mathcal{P}_E$. Inequalities can be used to compare the relative balance of core and periphery in the system and environment. Consider Figure \ref{fig:sym}, which considers the various possible outcomes. In the upper-right cases, system $S$ is more periphery-dominant than the environment $S_E$. In the diagonal cases, the relative balance of variety is the same between $S$ and $S_E$, i.e., there is symmetry between the system and environment. And in the lower-left cases, $S$ is more core-dominant than $S_E$. This is a useful exposition of the general regime. Given that $S$ is a regulator of $S_E$, it is useful to know if a homeostatic $S_E$ is regulated with a similarly homeostatic $S$, or if a largely unstable $S_E$ is regulated by a homeostatic $S$, etc. But, it is hard to assign relative value to these various cases because symmetry alone does not make a statement regarding the variety of outcomes.

\subsubsection{Blocking}

Outcome-based value judgements on the distribution of variety across $\mathcal{C}$ and $\mathcal{P}$ can be made by utilizing the Law of Requisite Variety to consider what $S$ is demonstrating between $t$ and $t'$ regarding its mechanism to block varieties in $S_E$, i.e., to decrease the lower bound in Equation \ref{eq:req}) and regulate outcomes. For example, is the variety in the environment's periphery $V_{\mathcal{P}_E}$ being treated with $V_\mathcal{P}$ or $V_\mathcal{C}$? One cannot know generally, but, one can deduce given conditions on variety. If $S$ is a stable regulator of $S_E$ and $V_{\mathcal{C}_E} > V_\mathcal{C}$, then one can deduce that the system must be partially addressing the variety in the environment's core with variety in the system's periphery\footnote{Note Equation \ref{eq:req} specifically concerns the variety of outputs in the system's core and periphery and the variety of inputs in the environment's core and periphery.}. Without making restrictive or unrealistic assumptions about the functional dependence of components in $S$, as traditional engineering practice does to use precepts of decomposition and recomposition, one can use precepts of core and periphery, defined over the component sets of $S$ and $S_E$, to model what aspects of an intelligent system are being used to block environmental variety in order to regulate outcomes.

\subsubsection{Abstraction Independence}

One may care about a subsystem of $S$. If one wants to know if a subsystem is in the core or periphery, one can just compare the subsystem to $\mathcal{C}$ or $\mathcal{P}$. However, one may find $S$ hard to model as a whole. But since the subsystem is a system, it can have its own core and periphery. Therefore, the use of core and periphery does not require observability of the entire system. Moreover, by considering the distribution of core and periphery across subsystems, it can be used to compare the interaction of subsystems within a system without making strong assumptions regarding independence. And, furthermore, it follows that modeling the core and periphery at the system, subsystem, and component levels of abstraction can identify how the core and periphery are distributed across a system. When combined with similar, stratified models of the environment's core and periphery, this provides a abstraction-independent means of modeling the relation between system and environment.

\subsubsection{Dynamics}

The core and periphery can be modeled over time. As such, the membership of elements (in the component sets) of a system can be traced as they move between the core and periphery. This provides a natural means of tracing adaptation in a system. Instead of facing the difficult task of comparing the self-similarity of components, subsystems, and their inter-relations at different points in time, modelers can simply demarcate the varying presence of residual complexity. In essence, one can model core and periphery growing or shrinking, and as such, address detailed questions regarding adaptation without traditional assumptions of component-level independence. E.g., if a large change occurs between $t$ and $t'$ in the environment, does the system change from $t'$ to $t''$ in response? Was the change in $S_E$ regulated by $S$, i.e., was $\min V_\mathcal{Z} = 0$ from $t$ to $t'$? Is there evidence of $S$ absorbing new varieties into its core from the periphery, i.e., is $\mathcal{C}^{t', t''} \cap \mathcal{P}^{t, t'}$ non-empty? Core and periphery supports complex and varied analysis into the dynamical nature of the relation between an intelligent system and its environment.

\subsection{Relevance}

Traditional precepts are well-established, widely applied, and writ large successful. It is important, then, to identify where specifically new precepts are needed. The core roughly corresponds to traditional engineering practice. While the components in the core may not be independent of each other, the core's stability suggests that their respective input-output relations are stable, and therefore can be subject to functional requirements. Consider the preceding example in DNNs. Firstly, those identifiable functions for passing information from layer to layer, etc., that are common across applications of DNNs can be associated with the core. Alternatively, the parameter values of a DNN and the data used to train it (if data is considered in scope) can be treated as parts of the periphery.

Having disambiguated the core of DNNs from the periphery, the traditional decomposition and recomposition precepts can be applied to the core. Whether passing information between layers or back-propagating error, functions of the core of DNNs have a mechanical, largely environment-independent and therefore universal character. In contrast, the same decomposition cannot be applied to the DNN's periphery. Various no free lunch theorems suggest that good training data and model parameters are not universal. While there are desirable, general properties of learned representations like linear separability, many such properties are already implicit in loss functions generally and therefore embedded into the core. This example in DNNs highlights that core and periphery are general precepts, and decomposition and recomposition are, in the main, precepts applicable to the core.

In this sense, the precepts of core and periphery reduce to traditional precepts of decomposition and recomposition for (simple) systems wherein input-output relations are easily attributable to outcomes. In such a case, a definition of intelligence as the property of a system is sufficient. Now, consider that outcomes are not easily attributable to input-output relations when boundaries are not well-defined. And then consider that coupling between systems, between subsystems, and between components tends to dissolve boundaries. In such cases, input-output relations cannot be easily attributed to outcomes, thus, engineering intelligence as a system property has insufficient scope to regulate outcomes, and therefore intelligence ought to be treated as a property of the relation between a system and its environment. To the extent that general intelligence is emblematic of the latter case, precepts of core and periphery are more relevant to engineering general intelligence than traditional precepts.


\section{Conclusion}

Whereas functional decomposition and recomposition are precepts for open systems, core and periphery are precepts for closed systems, i.e., for engineering intelligence as a property of the relation between system and environment. And whereas functional composition is associated writ large with stratification, hierarchies, and hierarchical engineering processes, the core and periphery are associated with a coarser disambiguation oriented towards characterizing the nature of inter-linkages created by intelligence. While closed systems may not appreciably exist in nature besides (perhaps) the universe, their emphasis here derives from a stated interest in formal precepts for engineering theory. Engineering---designing, building, and operating---AI-enabled systems needs to consider the necessity of new closed-system precepts for engineering AI towards stakeholders' desired outcomes.

Future work is needed to demonstrate and support practical value. First, the ability to empirically isolate system functions via core-periphery disambiguation should be evaluated on a system with general intelligence. Additionally, a longer-form, formal elaboration of core and periphery in terms of mathematical theorems and corollaries is needed. Similarly, a point-by-point comparison with traditional system engineering methods is merited. Lastly, determining core and periphery requires defining boundaries between systems, even if only temporarily. Additional research on the dynamics of boundaries between highly coupled systems is needed.

\bibliographystyle{splncs04}
\bibliography{ref}

\end{document}